\documentclass[letterpaper]{article} 
\usepackage{aaai25}  
\usepackage{times}  
\usepackage{helvet}  
\usepackage{courier}  
\usepackage[hyphens]{url}  
\usepackage{graphicx} 
\usepackage{amsfonts} 
\usepackage{booktabs}
\usepackage{multirow}
\usepackage{amsmath}
\usepackage{natbib}  
\usepackage{caption} 
\usepackage{algorithm}
\usepackage{algorithmic}
\usepackage{siunitx}

\usepackage{newfloat}
\usepackage{listings}
\DeclareCaptionStyle{ruled}{labelfont=normalfont,labelsep=colon,strut=off} 
\floatstyle{ruled}
\newfloat{listing}{tb}{lst}{}
\floatname{listing}{Listing}

\title{The Map of Misbelief: Tracing Intrinsic and Extrinsic Hallucinations \\Through Attention Patterns}
\author{
    Elyes Hajji,
    Aymen Bouguerra,
    Fabio Arnez
}
\affiliations{
    Université Paris-Saclay, CEA-List\\F-91120, Palaiseau, France\\
    name.lastname@cea.fr
}

\begin{document}

\maketitle

\begin{abstract}
Large Language Models (LLMs) are increasingly deployed in safety-critical domains, yet remain susceptible to hallucinations.
While prior works have proposed confidence representation methods for hallucination detection, most of these approaches rely on computationally expensive sampling strategies and often disregard the distinction between hallucination types.
In this work, we introduce a principled evaluation framework that differentiates between extrinsic and intrinsic hallucination categories and evaluates detection performance across a suite of curated benchmarks.
In addition, we leverage a recent attention-based uncertainty quantification algorithm and propose novel attention aggregation strategies that improve both interpretability and hallucination detection performance.
Our experimental findings reveal that sampling-based methods like Semantic Entropy are effective for detecting extrinsic hallucinations but generally fail on intrinsic ones. In contrast, our method, which aggregates attention over input tokens, is better suited for intrinsic hallucinations. These insights provide new directions for aligning detection strategies with the nature of hallucination and highlight attention as a rich signal for quantifying model uncertainty.

\end{abstract}

\section{Introduction}
\begin{figure}
    \centering
    \includegraphics[width=0.47\textwidth]{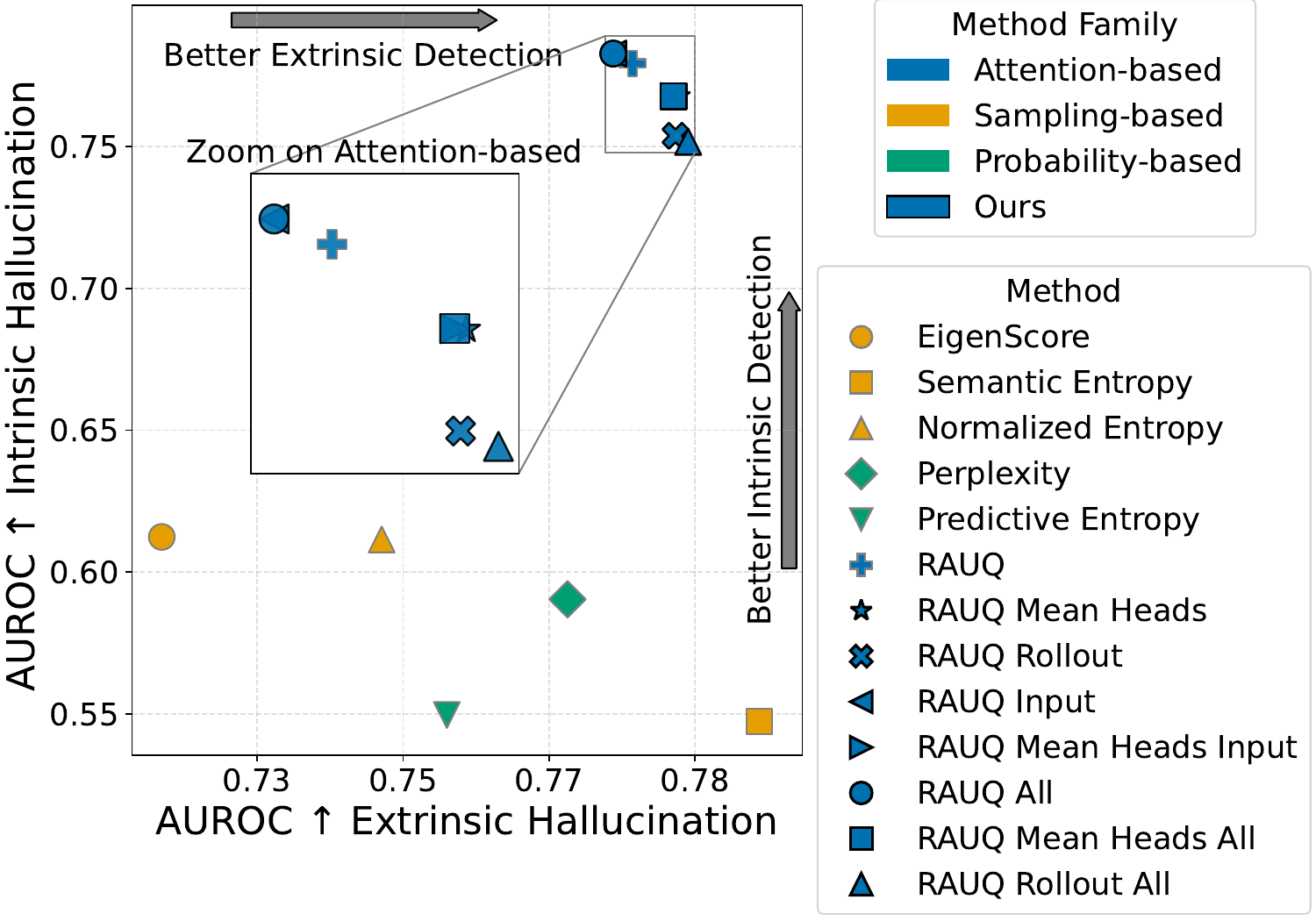}
    \caption{The Hallucination Map. AUROC scores for uncertainty estimation methods on extrinsic vs. intrinsic hallucination detection across all datasets. The proposed aggregation strategies in the RAUQ attention-based method outperform baselines and exhibit balanced performance across both hallucination types.}
    \label{fig:map}
\end{figure}
LLMs have demonstrated impressive capabilities across diverse natural language processing tasks \cite{10.1145/3744746} and are increasingly trusted in high‑stakes domains \cite{hung2023walkingtightropeevaluating}. Yet they remain prone to hallucinations \cite{Huang_2025, xu2025hallucinationinevitableinnatelimitation, Ji_2023}, i.e., outputs that are fluent but factually incorrect or unsupported. At their core, LLMs rely on statistical patterns and probabilistic inference to complete text sequences, without any built‑in sense of their own knowledge boundaries. Instead, they are typically optimized to maximize helpfulness and confidence, which can lead them to fabricate plausible‑sounding responses when information is lacking.

The term hallucination in the context of language models is broad and commonly used loosely, which makes precise evaluation and mitigation challenging. To address this, \citet{Ji_2023} proposed a more structured taxonomy that distinguishes between two main types of hallucinations. Extrinsic hallucinations refer to model outputs that are not supported by the model’s training data, reflecting its inability to differentiate between known and unknown information. On the other hand, intrinsic hallucinations occur when the generated content contradicts the input context itself, such as summarizing facts inaccurately or fabricating details not found in the input source.

Hallucinations in language models are often identified by assessing the model's confidence in its outputs, for example, by quantifying the uncertainty in its predictions or generated responses.
Traditionally, uncertainty is split into epistemic uncertainty, the model’s lack of knowledge, which can be reduced with more data, and aleatoric uncertainty, irreducible noise inherent in the data. \citet{kirchhof2025positionuncertaintyquantificationneeds} challenge this simple categorization, demonstrating that both forms can shift fluidly depending on context.
To capture uncertainty stemming from ambiguous or incomplete user queries rather than model limitations, they introduce underspecification uncertainty. This type of uncertainty can be associated with intrinsic hallucination. For example, when inputs are underspecified, the model is forced to generate information that lacks sufficient grounding, thereby increasing the chance of producing hallucinated outputs.

Existing methods for hallucination detection typically fall into three categories: sampling-based approaches, probability-based scores, and techniques that leverage internal model representations. Sampling-based methods \cite{kuhn2023semanticuncertaintylinguisticinvariances, chen2024insidellmsinternalstates, manakul2023selfcheckgptzeroresourceblackboxhallucination} can be effective but are often prohibitively expensive for practical use. Probability-based scores, such as perplexity, are lightweight but frequently fail to capture deeper inconsistencies. Some recent approaches involve training separate detectors, though these models often struggle to generalize across tasks. A promising direction involves exploiting internal model representations \cite{quevedo2024detectinghallucinationslargelanguage, kossen2024semanticentropyprobesrobust, vazhentsev2025uncertaintyawareattentionheadsefficient}, which often carry latent indicators of uncertainty, offering a potential pathway to detect when a language model is operating beyond its safe knowledge limits. Nevertheless, prior work typically does not differentiate between types of hallucinations. Distinctions are typically framed as a contrast between factuality and faithfulness, an approach that has been proven both inadequate and misleading \cite{bang2025hallulensllmhallucinationbenchmark}.
This work builds on more precise typologies, recognizing the fundamentally different nature of intrinsic and extrinsic hallucinations, and further leverages attention patterns and aggregation techniques for lightweight hallucination detection. Our key contributions are threefold:

\begin{itemize}
    \item We propose novel attention aggregation strategies for uncertainty quantification in hallucination detection, with some variants improving extrinsic detection and others enhancing intrinsic detection, thereby advancing both accuracy and interpretability over prior attention-guided methods.
    \item We develop a structured benchmarking protocol that explicitly distinguishes between intrinsic and extrinsic hallucinations, leveraging curated datasets and a comprehensive suite of metrics (AUROC, AURAC, PRR). This framework ensures fair and type‑specific assessment of hallucination detection techniques.
    \item We perform extensive experiments on six open-source LLMs and reveal that current leading sampling-based methods often fail to detect certain types of hallucinations as seen in Figure \ref{fig:map}, particularly intrinsic errors caused by underspecification, highlighting the need for more robust detection approaches.
\end{itemize}

\section{Related Work}
A wide range of uncertainty-based techniques has been proposed for hallucination detection, differing in how they estimate or model confidence in generated outputs.

Sampling-based methods have emerged as some of the most prominent techniques for hallucination detection, demonstrating strong performance. For instance, SelfCheckGPT \cite{manakul2023selfcheckgptzeroresourceblackboxhallucination} assesses the consistency of a generated sentence by comparing it to multiple stochastically sampled responses, using an LLM itself to judge whether each sample supports the original claim. Semantic Entropy \cite{kuhn2023semanticuncertaintylinguisticinvariances} builds on a similar idea of semantic agreement but formalizes it through entropy computed over clusters of meaning in the sampled outputs. EigenScore \cite{chen2024insidellmsinternalstates} also leverages semantic information, but operates in the embedding space by measuring differential entropy to capture variation in the semantic representations of the sampled responses.

While early sampling‐based methods generated multiple outputs under uniform decoding parameters, subsequent work has introduced more structured diversity mechanisms. \citet{cecere2025montecarlotemperaturerobust} propose Temperature Monte Carlo Sampling, varying the temperature across samples. \citet{balabanov2025uncertainty} form an ensemble by training multiple LoRA adapters with different initializations, and \citet{arteaga2024hallucination} further reduce cost by combining a single LoRA adapter with a BatchEnsemble technique. \citet{zhang2024vluncertaintydetectinghallucinationlarge} achieve input‐level variation through semantically equivalent prompt reformulations. Despite their effectiveness, these sampling‐based approaches remain prohibitively expensive for real‐time deployment.

An alternative line of research trains dedicated detectors on internal model features to enable single‑pass uncertainty estimation. \citet{kossen2024semanticentropyprobesrobust} employ lightweight linear probes to predict entropy from hidden states, while \citet{quevedo2024detectinghallucinationslargelanguage} extract token-probability features from model outputs, which are then used to train a lightweight classifier for hallucination detection. Although these methods perform well on in‑domain benchmarks, they typically depend on dataset‑specific supervision, which undermines their generalizability.

Another class of approaches leverages attention patterns within language models to quantify uncertainty. \citet{zhang2023enhancinguncertaintybasedhallucinationdetection} propose a method that propagates uncertainty across generation steps using attention dynamics. \citet{vazhentsev2024unconditionaltruthfulnesslearningconditional} and \citet{chuang2024lookbacklensdetectingmitigating} also extract attention-based features, which are fed into trained detectors to assess the reliability of generated outputs.
More recently, \citet{vazhentsev2025uncertaintyawareattentionheadsefficient} introduced RAUQ, a lightweight method that selects specific attention scores based on observation and uses them as weights to propagate uncertainty across decoding steps, offering a computationally efficient technique that beats state-of-the-art methods.

Despite the proliferation of hallucination detection techniques, nearly all prior evaluations rely on non‑curated benchmarks originally designed to assess overall LLM capabilities rather than targeted hallucination detection, which prevents a clear distinction between different hallucination types. \citet{bang2025hallulensllmhallucinationbenchmark} address this gap with HalluLens, the first benchmark specifically designed to target and evaluate different types of hallucination separately.

\section{Preliminaries}

\subsection{Hallucination Types}
\paragraph{Extrinsic Hallucination.}Arises when a model generates information that is not grounded in its training data. Given that modern LLMs are typically pretrained on large corpora, including the full contents of Wikipedia, any factual query about covered material (excluding events after the model’s knowledge cutoff) should elicit a correct response. When the model instead fabricates or recalls incorrect facts, it is exhibiting an extrinsic hallucination, revealing a failure to recognize or retrieve knowledge it ostensibly possesses.
\paragraph{Intrinsic Hallucination.}They occur when a model’s output contradicts or oversteps the given input context. For example, if the context itself contains information that conflicts with real‑world facts, the model must adhere to that context, even if it “knows” the real fact; otherwise, it hallucinates intrinsically. In another scenario, the context may omit critical details needed to complete a task, yet the model still invents the missing information and responds. In both cases, the model fails to ground its output in the input, 
and is prone to underspecification uncertainty due to the ambiguous or insufficient context, which highlights the connection between intrinsic hallucination and this type of uncertainty.

Intrinsic and extrinsic hallucinations reveal the limitations of the standard separation between \emph{faithfulness} versus \emph{factuality}. Under that framing, any output deemed factually correct is automatically considered faithful, even if it directly contradicts the input context. For instance, 
For example, if the input context states “Google was founded in 2001” but the model's response contains “Google was founded in 1998” the answer is factually correct but unfaithful to the given context which is an intrinsic hallucination.
 
This example reveals how the faithfulness/factuality split conflates the \emph{source} of an error (misalignment with input versus ignorance of world knowledge) with the \emph{evaluation criterion} (consistency versus correctness). By distinguishing intrinsic hallucinations (input‐contradicting) from extrinsic hallucinations (knowledge‐gap), our taxonomy separates these dimensions and supports more targeted detection strategies.

\subsection{RAUQ: Recurrent Attention-based Uncertainty Quantification}

\citet{vazhentsev2025uncertaintyawareattentionheadsefficient} first observed that, across multiple qualitative examples, attention weights from each new token back to its immediate predecessor tend to drop whenever the model is on track to produce an incorrect or hallucinatory continuation. They further show that this attention drop consistently appears only in a small subset of attention heads, termed \emph{uncertainty-aware heads}, while the remaining heads maintain stable, low attention patterns (see Figure 1 of \cite{vazhentsev2025uncertaintyawareattentionheadsefficient}). This phenomenon suggests that attention dynamics themselves carry a strong signal of local model confidence. Accordingly, for each layer \(\ell\), one selects its uncertainty‑aware head \(h_\ell\) by finding, over all heads, the head whose average predecessor-to-current-token attention is maximal:
{\footnotesize
\begin{equation}
\label{rauq_agg}
    h_\ell = \underset{h=1\dots H}{\arg\max}\ 
      \Bigl(\frac{1}{T-1}\sum_{t=2}^{T} \mathcal{A}^{\ell, h}_{t, t-1}\Bigr).
\end{equation}}

Since language models generate text strictly sequentially, each emitted token \(y_t\) becomes part of the fixed context for all subsequent decoding steps, any error in \(y_t\) can compound downstream. Therefore, the model’s confidence in each new token naturally depends not only on its own base probability but also on the accumulated certainty of the preceding tokens. This insight underlies the uncertainty‑propagation mechanism introduced in \cite{zhang2023enhancinguncertaintybasedhallucinationdetection}.

Building on these observations, the RAUQ algorithm treats generation as a sequential process in which each token’s confidence is a blend of the model’s base probability \(p_t\) for that token and the propagated confidence carried over from the previous step. Concretely, at layer \(\ell\), RAUQ updates the propagated confidence \(c_\ell(y_t)\) according to:
{\footnotesize
\begin{equation}
\label{rauq_conf}
    c_\ell(y_t)\;\gets\;
          \alpha\,p_t
          \;+\;(1-\alpha)\,\mathcal{A}^{\ell, h_\ell}_{t, t-1}\,c_\ell(y_{t-1})
\end{equation}}
where \(\alpha\in[0,1]\) controls the trade‑off between the token’s own probability and the propagated signal, and \(\mathcal{A}\) denotes the attention map.

In the RAUQ algorithm, for each layer \(\ell\) one performs the following steps:
\begin{enumerate}
    \item Based on observation, select the uncertainty‑aware head by aggregating per‑token attention into a head‑level score (Eq.~\ref{rauq_agg}).
    \item Compute each token’s propagated confidence using that head (Eq.~\ref{rauq_conf}).
    \item Aggregate these token‑level confidences into a single uncertainty score for the layer by taking the negative mean logarithm:
    {\footnotesize
    \begin{equation}
        u_\ell = -\frac{1}{T}\sum_{t=1}^T \log c_\ell(y_t)
    \end{equation}}
\end{enumerate}
Finally, the overall uncertainty of the LLM’s output is defined as the maximum layer‑wise uncertainty.

\section{Method}
\citet{abnar-zuidema-2020-quantifying} observed that raw attention weights in Transformer models tend to become less interpretable in deeper layers, often becoming uniformly distributed and weakly correlated with other importance measures. These findings suggest that naive attention extraction may obscure meaningful patterns. Motivated by this, we explore alternative aggregation strategies and adopt the uncertainty propagation mechanism from the RAUQ algorithm to build a more robust attention-based hallucination detection method.
We investigate three token aggregation strategies for attention weights, each differing in the span of tokens considered:
\begin{itemize}
  \item \textbf{Previous‐Token Aggregation (Original RAUQ baseline):}
    {\footnotesize
    \begin{equation}
        a_t^{\ell,h} = \mathcal{A}^{\ell,h}_{t, t-1}
    \end{equation}}
          
  \item \textbf{All‐Past‐Tokens Aggregation:}
    {\footnotesize
    \begin{equation}
      a_t^{\ell,h} = \frac{1}{m + (t-1)} \sum_{j=1}^{m+t-1} \mathcal{A}^{\ell,h}_{t,j}
    \end{equation}}
    where \(m\) is the number of input tokens.

    \citet{vazhentsev2025uncertaintyawareattentionheadsefficient} performed a limited ablation by comparing attention from the current token t to a single prior token at various offsets $(t-1,t-2,\dots,t-6)$. In contrast, our All‑Past‑Tokens variant computes the average attention from $t$ to every preceding token (including the full input sequence). Since the model’s next‐token prediction inherently conditions on its entire prior context, this aggregation more faithfully captures the cumulative uncertainty propagated through all previously generated tokens.
  \item \textbf{Input‐Tokens Aggregation:}
    {\footnotesize
    \begin{equation}
      a_t^{\ell,h} = \frac{1}{m} \sum_{i=1}^{m} \mathcal{A}^{\ell,h}_{t,i}
    \end{equation}}
    In tasks such as extractive question answering, summarization, or retrieval‑augmented generation, tasks where the model’s output must be entirely grounded in the source, strong attention to the input is essential. If, when generating a token, the model does not attend back to the input context, it likely indicates reliance on internal priors rather than evidence from the input, and thus a higher risk of hallucination. By averaging attention over all input tokens, this variant directly measures how much the model leans on the provided context at each decoding step.
\end{itemize}

For each of the above token aggregation variants, we evaluate three attention head‐aggregation modes:

\begin{itemize}
  \item \textbf{Original RAUQ Head Selection:}  
    Select the head \(h_\ell\) in each layer by maximal mean (Eq. 1).
    {\footnotesize
    \begin{equation}
         a^\ell_t = a^{\ell, h_\ell}_t
    \end{equation}}

  \item \textbf{Mean Across Heads:}  
    Replace head selection by averaging across all \(H\) heads:
    {\footnotesize
    \begin{equation}
        a_t^{\ell} = \frac{1}{H}\sum_{h=1}^{H} a_t^{\ell,h}
    \end{equation}}
  The identification of a single “uncertainty‑aware” head is purely observational, and may overlook useful signals in the remaining heads. Averaging across all heads tests whether a collective summary of attention patterns provides a more reliable and stable measure of model uncertainty than selecting one head alone.

    \item \textbf{Attention Rollout} \cite{abnar-zuidema-2020-quantifying}: Recursively multiply attention matrices (averaged over heads) across layers to estimate how attention compounds from input tokens to output tokens throughout the model:
    {\footnotesize
    \begin{equation} 
        R^{(1)} = \tilde{\omega}^{(1)}, \quad
        R^{(\ell)} = \tilde{\omega}^{(\ell)} \cdot R^{(\ell - 1)} \quad \text{for } \ell > 1
    \end{equation}}
    where $\tilde{\omega}^{(\ell)} = \frac{1}{2}\left( \omega^{(\ell)} + I \right)$, $\omega^{(\ell)} = \frac{1}{H}\sum_{h=1}^{H} \mathcal{A}^{\ell,h}$ and and \(I\) is the identity matrix to simulate the residual connection.\\ We evaluate this aggregation method using both attention to the immediately preceding token ($a^\ell_t = R^\ell_{t, t-1}$) and the average attention across all previously generated tokens ($a_t^{\ell} = \frac{1}{m + (t-1)} \sum_{j=1}^{m+t-1} R^{\ell}_{t,j}$).\\

\end{itemize}

\section{Experiments}

\subsection{Datasets}

We propose an evaluation framework on the question-answering (QA) task by assessing the extrinsic and intrinsic hallucination detection capabilities in LLMs, as described below.

For \textit{extrinsic} hallucination assessment, we employ the HalluLens benchmark \cite{bang2025hallulensllmhallucinationbenchmark} using the authors' public repository to generate its examples with Llama‑3.1‑70B‑Instruct.

This process yields two complementary subsets designed to test \emph{extrinsic} hallucination: 1) \textbf{PreciseWikiQA:} Curated open‑domain questions derived from Wikipedia, assessing the model’s ability to retrieve and ground answers in its pretraining knowledge; failures here indicate outputs unsupported by the training corpus. 2) \textbf{NonExistentRefusal‑MixedEntities:} Prompts requesting details about never‑existent entities, evaluating whether the model can recognize its knowledge limits and refuse or express uncertainty; success here demonstrates resistance to fabricating unsupported content.

We assess \emph{intrinsic} hallucination using the \textbf{SQuAD‑v2} dataset \cite{rajpurkar2018knowdontknowunanswerable}, an extractive QA benchmark comprising passages paired with questions, 50\% of which are unanswerable (i.e., no valid answer span exists in the context). Unanswerable instances serve as a proxy for intrinsic hallucinations, since the correct response must be derived from the input. We also include two subsets of the \textbf{FaithEval} benchmark \cite{ming2025faithevallanguagemodelstay}: 1) The \textbf{counterfactual} subset introduces passages that contradict real-world knowledge, challenging the model to rely exclusively on the provided context rather than fallback on prior knowledge, a suitable test for intrinsic hallucination detection. 2) The \textbf{inconsistent} subset, on the other hand, contains passages with internal contradictions, making all associated questions unanswerable and thus directly assessing the model's ability to identify inconsistencies within the input itself.
These datasets target intrinsic hallucination, and the inability to detect such inconsistencies may also reflect \textit{underspecification} uncertainty defined by \cite{kirchhof2025positionuncertaintyquantificationneeds}.

We additionally use \textbf{CoQA} \cite{reddy2019coqaconversationalquestionanswering}, an extractive question answering dataset, as well as \textbf{TriviaQA} \cite{joshi2017triviaqalargescaledistantly} and \textbf{NQ-Open} \cite{47761}, which are open-domain QA datasets, as they are commonly used in prior work on hallucination detection for comparative evaluation.
\subsection{Models}
We evaluate our approach on a range of open-source instruction-tuned LLMs to ensure accessibility to internal attention maps and token probabilities. The models include LLaMA-3.1-8B-Instruct, Mistral-7B-Instruct-v0.3, Falcon-10B-Instruct, Gemma-2-9B-It, Qwen-2.5-7B-Instruct and Mistral-Nemo-Instruct-2407. Each method is tested on every model across all presented datasets to ensure robust and comprehensive comparisons.

\subsection{Baselines}
We employ sampling-based uncertainty estimation methods by generating multiple responses from the model. Specifically, we sample 10 outputs using nucleus sampling with a high temperature ($temperature = 1.0$) to encourage response diversity, and one deterministic response using greedy decoding with low temperature ($temperature = 0.1$) to approximate a confident prediction. These generations are then used to compute uncertainty using three techniques: \textbf{Semantic Entropy} \cite{kuhn2023semanticuncertaintylinguisticinvariances}, \textbf{EigenScore} \cite{chen2024insidellmsinternalstates}, and \textbf{Length Normalized Entropy (Normalized Entropy)} \cite{malinin2021uncertainty}.\\
We also evaluate two commonly used confidence-based baselines: \textbf{Perplexity}, which measures the inverse probability of the generated sequence, and \textbf{Average Predictive Entropy}, which computes the average entropy of the model's token-level output distributions. These methods rely solely on token probabilities and serve as strong baselines for assessing uncertainty.

\subsection{Evaluation Metrics}
\textbf{Response Evaluation:}
To assess the correctness of model responses, we use \textbf{AlignScore} \cite{zha2023alignscoreevaluatingfactualconsistency}, as it provides a more fair and context-aware judgment of output quality compared to purely lexical metrics. 

It is used as the primary correctness metric across all datasets, except for \textbf{NonExistentRefusal‑MixedEntities}, which was manually annotated: responses were assigned a score of 1 if the model explicitly expressed lack of knowledge about the entity, and 0 if it provided fabricated information.\\
\textbf{Hallucination Detection Evaluation:}
To evaluate the quality of uncertainty estimates for hallucination detection, we adopt the following metrics:

\begin{itemize}
    \item \textbf{Area Under the Receiver Operating Characteristic curve (AUROC):} It quantifies how well the uncertainty scores separate hallucinated from non-hallucinated outputs.

    \item \textbf{Area Under the Rejection–Accuracy Curve (AURAC)}: 
    This metric was introduced by \citet{kuhn2023semanticuncertaintylinguisticinvariances} and it summarizes how well a model can improve accuracy by rejecting predictions with high uncertainty. The rejection–accuracy curve plots the accuracy of accepted predictions as a function of the rejection rate. AURAC captures the area under this curve.

    \item \textbf{Prediction Rejection Ratio (PRR)} \cite{malinin2021uncertainty}: 
    This metric evaluates the effectiveness of the model's rejection policy relative to an oracle and a random baseline. Prediction rejection curves plot the average answer quality (e.g., AlignScore) after replacing rejected predictions with oracle responses, as a function of the rejection rate. PRR normalizes the performance between the random and oracle strategies, providing an interpretable score.
\end{itemize}

For methods that do not rely on sampling, all necessary model outputs (token probabilities and attention maps) are obtained from the confidently generated response. This same response is also used to evaluate correctness against the gold reference using AlignScore.
For the uncertainty propagation methods (RAUQ and our variants), the hyperparameter~$\alpha$ was tuned via grid search over the range $[0.1, 0.2, \dots, 0.9]$ to maximize AUROC on the curated benchmarks (HalluLens, SQuAD-v2, and FaithEval).

\section{Results \& Discussion}
\begin{figure}[t!]
    \centering
    \includegraphics[width=0.47\textwidth]{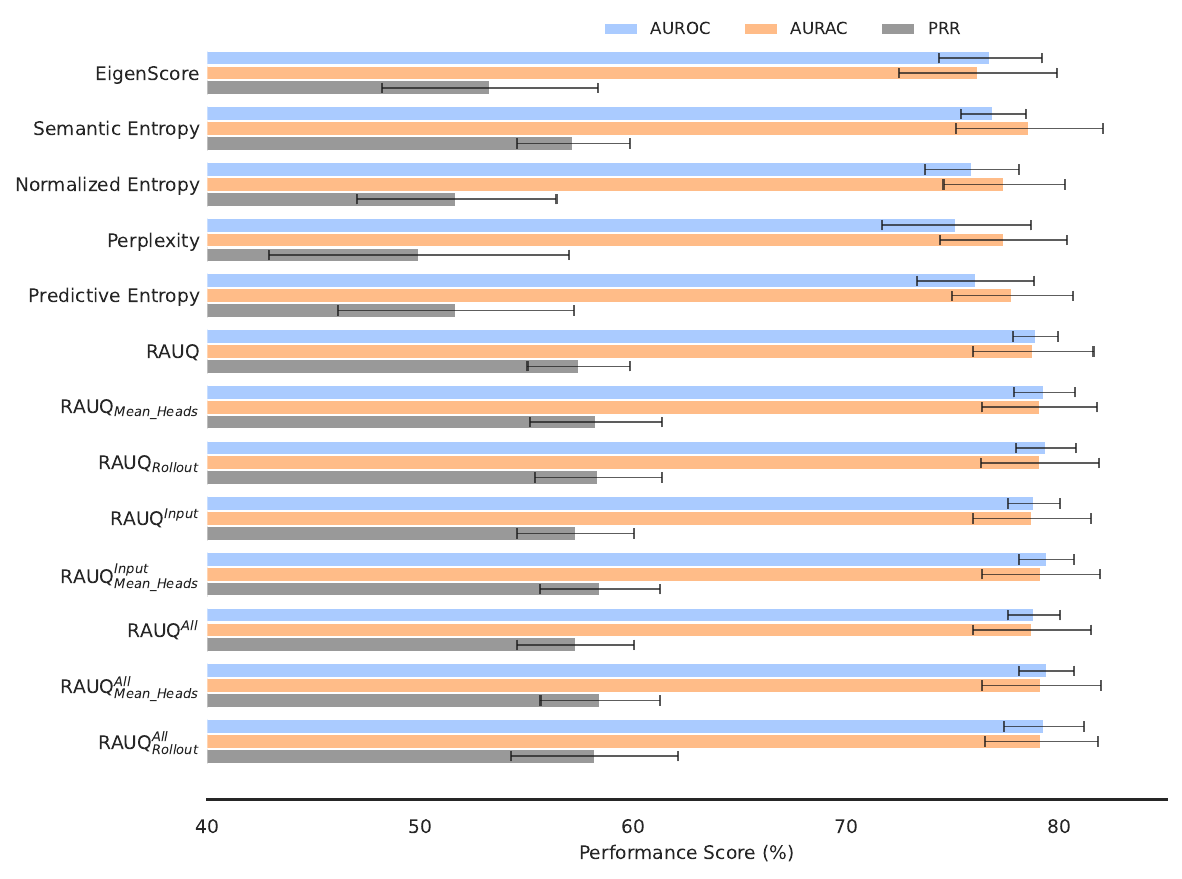}
    \caption{Comparison of AUROC$\uparrow$, AURAC$\uparrow$, and PRR$\uparrow$ scores for all hallucination detection methods, computed over the concatenation of all datasets. Error bars indicate $\pm$ one standard deviation across models. The axis for each metric is colored to match its bars, indicating the corresponding scale. Dashed lines indicate the performance of the RAUQ baseline. This figure provides a comprehensive overview of each method’s overall detection performance. Some RAUQ variants, particularly the Rollout and Mean‑Heads aggregations, consistently achieve the highest performance overall.}
    \label{fig:all}
\end{figure}
\begin{figure*}[t!]
    \centering
    \includegraphics[width=1\textwidth]{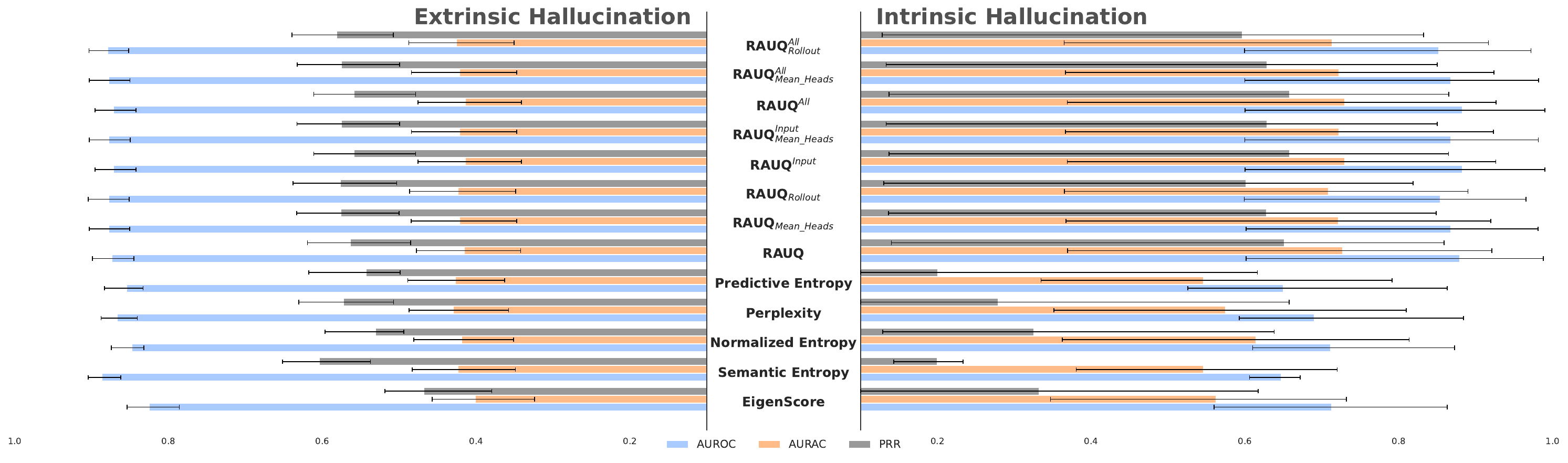}
    \caption{Comparison of hallucination detection methods across \textit{extrinsic} and \textit{intrinsic} hallucination benchmarks using AUROC$\uparrow$, AURAC$\uparrow$, and PRR$\uparrow$. Error bars indicate the 95\% confidence intervals. The axis for each metric is colored to match its bars, indicating the corresponding scale. Dashed lines indicate the performance of the RAUQ baseline. While Semantic Entropy performs well on extrinsic hallucinations, attention-based variants aggregating over input or all tokens consistently achieve higher performance on intrinsic hallucinations. The butterfly split highlights divergent performance trends depending on hallucination type.}
    \label{fig: ext_int}
\end{figure*}
We evaluate all methods across three settings: overall (all datasets combined), extrinsic hallucination detection, and intrinsic hallucination detection. Figure~\ref{fig:all} presents a comparison of AUROC, AURAC, and PRR scores across all datasets, offering a global performance summary. Figure~\ref{fig: ext_int} complements this by contrasting method performance on extrinsic versus intrinsic hallucinations, highlighting divergent trends depending on the hallucination type.

Note that, to offer a more data‑sensitive comparison, we concatenate all instances from the evaluated benchmarks and compute a single set of performance metrics (AUROC, AURAC, and PRR) for each method. This contrasts with the more common practice of reporting average metric values across datasets, which can disproportionately amplify the influence of smaller benchmarks. Our approach yields a more representative assessment of each method’s detection performance over the full range of hallucination examples.

\subsection{Overall Performance}
As shown in Figure~\ref{fig:all}, RAUQ with our attention‑aggregation variants achieve the highest AUROC when evaluated over the combined set of all datasets. In particular, the \emph{Rollout}, \emph{Rollout All}, \emph{Mean‑Heads All}, and \emph{Mean‑Heads Input} variants outperform the original RAUQ baseline. The near‑identical performance of the \emph{All‑Past‑Tokens} and \emph{Input‑Tokens} variants arises from generally low attention magnitudes, suggesting they could be unified into a single averaging strategy.
\subsection{Extrinsic Hallucination}
On the extrinsic benchmarks (PreciseWikiQA and NonExistentRefusal-MixedEntities), Figure \ref{fig: ext_int} shows that the \emph{Semantic Entropy} sampling method attains the highest performance, surpassing both attention‑based and probability‑based approaches. This indicates that, when factual information is missing, the model’s outputs become highly variable, elevating semantic entropy and making it a strong indicator of extrinsic hallucinations.
\begin{figure}[t!]
    \centering
    \includegraphics[width=0.40\textwidth]{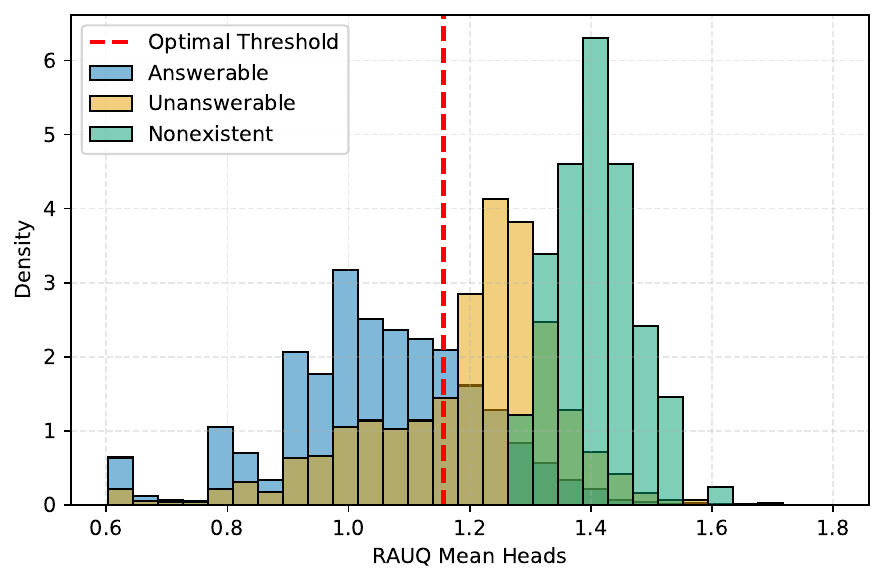}
    \caption{Normalized histograms showing the distribution of RAUQ Mean Heads scores for Mistral‑7B across SQuAD v2 Answerable (blue), Unanswerable (orange), and NonExistentRefusal-MixedEntities (grey). The red dashed line indicates the optimal threshold, obtained by maximizing the G-mean (TPR and TNR) across all datasets.}
    \label{fig:dist}
\end{figure}

Figure \ref{fig:dist} presents histogram-based distributions of RAUQ Mean Heads scores for Mistral‑7B on three types of examples: Answerable and Unanswerable questions from SQuAD-v2, and Nonexistent queries from our extrinsic hallucination benchmark. The optimal uncertainty threshold, marked with a red dashed line, was computed by maximizing the geometric mean (G-mean) of true positive and true negative rates over all datasets combined.
The visualization reveals a clear separation: Nonexistent queries exhibit high scores, reflecting consistent uncertainty in the model’s responses. In contrast, answerable queries cluster at lower RAUQ values, suggesting the model is confident when its answers are grounded in the context, supported by a high AlignScore of 0.86.

The RAUQ Mean Heads score for unanswerable questions from SQuAD-v2 falls between that of clearly answerable and clearly unanswerable (Nonexistent) examples, indicating that the model exhibits increased uncertainty in response to underspecified inputs. However, despite this internal signal, the model still produces incorrect answers, as reflected by an AlignScore of just 0.27, highlighting the gap between internal uncertainty and surface-level behavior.
\subsection{Intrinsic Hallucination}
Conversely, as shown in Figure \ref{fig: ext_int}, sampling‑based methods (including semantic entropy) perform poorly on intrinsic benchmarks (SQuAD‑v2-unanswerable, FaithEval‑counterfactual, FaithEval‑inconsistent), a finding that indicates contradictory inputs produce consistently structured, rather than diverse, outputs. In this setting, attention-based variants that aggregate over the input or all past tokens deliver the best detection performance, an intuitive result, as they rely on how much attention the model allocates to the provided context during generation.

While RAUQ$_{Mean\_Heads}$ scores indicate that Mistral exhibits uncertainty on SQuAD-v2 unanswerable questions in comparison across different datasets, as shown in Figure \ref{fig:dist}, this signal becomes less distinguishable when viewed within the dataset itself. As a result, performance for intrinsic hallucination drops, contributing to the observed variance across models in Figure \ref{fig: ext_int}.

\subsection{Discussion}
\begin{table}[t!]
\setlength{\tabcolsep}{1mm}
\begin{tabular}{llll}
\toprule
Method                     & AUROC($\uparrow$)           & AURAC($\uparrow$)           & PRR($\uparrow$)            \\ \midrule
EigenScore            & 0.4687          & 0.2860           & -0.0253        \\
Semantic Entropy      & 0.4692          & 0.3303          & 0.0025         \\
Normalized Entropy    & 0.5054          & 0.3568          & 0.0471         \\
Perplexity            & 0.5036          & 0.3599          & 0.0306         \\
Predictive Entropy    & 0.5097          & 0.3617          & 0.0395         \\
RAUQ                  & \underline{0.7092}          & 0.4567          & 0.4120          \\ \midrule
RAUQ$_{Mean\_Heads}$       & 0.6959          & 0.4522          & 0.3916         \\
RAUQ$_{Rollout}$         & 0.6852          & 0.4491          & 0.3707         \\
RAUQ$^{Input}$            & \textbf{0.7222} & \textbf{0.4669} & \textbf{0.4370} \\
RAUQ$_{Mean\_Heads}^{Input}$ & 0.7025          & 0.4600            & 0.4038         \\
RAUQ$^{All}$              & \textbf{0.7222} & \underline{0.4668}          & \underline{0.4369}         \\
RAUQ$_{Mean\_Heads}^{All}$   & 0.7025          & 0.4601          & 0.4039         \\
RAUQ$_{Rollout}^{All}$      & 0.6822          & 0.4542          & 0.3697         \\ \midrule \midrule
AlignScore Avg.    & \multicolumn{3}{c}{0.388}   \\ \bottomrule
\end{tabular}
\caption{Performance metrics (AUROC, AURAC, and PRR) for Qwen-2.5-7B-Instruct on the FaithEval‑inconsistent‑v1.0 dataset. Best values are shown in bold, and second-best values are underlined.}
\label{tab:qwen}
\end{table}

Our results demonstrate that uncertainty propagation via attention is a robust and computationally efficient strategy for detecting both extrinsic and intrinsic hallucinations. By leveraging a single pass through the model’s attention maps, these methods reduce computational cost by an order of magnitude compared to state-of-the-art approaches such as Semantic Entropy and EigenScore, which require multiple forward passes.

Moreover, the largest performance gains over baselines occur on the most challenging intrinsic benchmarks (FaithEval‑inconsistent and the unanswerable portion of SQuAD‑v2), where RAUQ variants that aggregate attention over the input or across all tokens substantially outperform other methods, as seen in Table \ref{tab:qwen}. This suggests that attention‐driven uncertainty estimation is particularly sensitive to underspecification uncertainty, where the model must recognize contradictions or missing information in the input. These findings motivate further exploration of attention dynamics as a proxy for underspecification uncertainties in LLMs.

While our RAUQ variants match or exceed baseline performance overall, Semantic Entropy remains the strongest detector of extrinsic hallucinations, reflecting its ability to capture response variability when factual information is absent. In practice, attention‑based methods deliver robust detection across both extrinsic and intrinsic hallucinations, while sampling‐based entropy methods excel when assessing factual grounding on open‐domain queries.

To complement the aggregate results shown earlier, Table \ref{tab:models} presents the full performance breakdown of all uncertainty estimation methods across the six tested open-source models. As before, AUROC, AURAC, and PRR are computed on the union of all datasets. This breakdown enables closer inspection of how our proposed methods generalize across model architectures.

While our evaluation framework has been applied to a variety of models and curated benchmarks, its effectiveness in complex reasoning tasks and real-world scenarios remains largely unexplored. Future research should prioritize examining whether LLMs' internal representations and attention patterns remain reliable in expressing confidence and enable dealing with the intricacies of more sophisticated tasks.

\newcommand{\metricblock}[3]{\begin{tabular}{@{}c@{}}#1\\#2\\#3\end{tabular}}
\newcommand{\metricblockbold}[3]{\textbf{\begin{tabular}{@{}c@{}}#1\\#2\\#3\end{tabular}}}

\begin{table}[t!]

\label{tab:fused_final_aligned}
\centering
\small
\setlength{\tabcolsep}{2pt} 
\sisetup{table-format=1.4}
\setlength{\tabcolsep}{2pt} 
\sisetup{table-format=1.4}
\begin{tabular}{@{} m{1.4cm} c c c c c c @{}}
\toprule
\textbf{Method} & \textbf{LLaMA} & \textbf{Mistral} & \textbf{M-Nemo} & \textbf{Falcon} & \textbf{Qwen} & \textbf{Gemma} \\
\midrule
ES      & \metricblock{0.7868}{0.7817}{0.5671} & \metricblock{0.7915}{0.7416}{0.5908} & \metricblock{0.7874}{0.8263}{0.5726} & \metricblock{0.7597}{0.7243}{0.5162} & \metricblock{0.7555}{0.7208}{0.5065} & \metricblock{0.7231}{0.7745}{0.4422} \\
\addlinespace
Sem. Ent. & \metricblock{0.7711}{0.7960}{0.5879} & \metricblock{0.7588}{0.7318}{0.5669} & \metricblock{0.7966}{0.8351}{0.6210} & \metricblock{0.7681}{0.7537}{0.5586} & \metricblock{0.7722}{0.7854}{0.5586} & \metricblock{0.7463}{0.8121}{0.5380} \\
\addlinespace
Norm. Ent. & \metricblock{0.7699}{0.7799}{0.5336} & \metricblock{0.7778}{0.7368}{0.5594} & \metricblock{0.7663}{0.8250}{0.5306} & \metricblock{0.7465}{0.7461}{0.4885} & \metricblock{0.7769}{0.7787}{0.5623} & \metricblock{0.7156}{0.7762}{0.4270} \\
\addlinespace
PPL & \metricblock{0.7583}{0.7740}{0.5105} & \metricblock{0.7627}{0.7378}{0.5259} & \metricblock{0.7721}{0.8284}{0.5433} & \metricblock{0.7560}{0.7516}{0.5059} & \metricblock{0.7837}{0.7910}{0.5626} & \metricblock{0.6762}{0.7587}{0.3477} \\
\addlinespace
Pred. Ent. & \metricblock{0.7772}{0.7856}{0.5476} & \metricblock{0.7720}{0.7408}{0.5444} & \metricblock{0.7708}{0.8275}{0.5397} & \metricblock{0.7572}{0.7499}{0.5072} & \metricblock{0.7842}{0.7906}{0.5625} & \metricblock{0.7018}{0.7720}{0.3981} \\
\addlinespace
RAUQ & \metricblock{0.7939}{0.7816}{0.5818} & \metricblock{0.7999}{0.7443}{0.6068} & \metricblock{0.7900}{0.8280}{0.5772} & \metricblock{0.7963}{0.7676}{0.5882} & \metricblock{0.7842}{0.7875}{0.5615} & \textbf{\metricblock{0.7679}{0.8164}{0.5298}} \\
\addlinespace
\midrule
RAUQ$_{\text{M-H}}$ & \metricblock{0.8008}{0.7872}{0.5953} & \metricblock{0.8035}{0.7478}{0.6133} & \metricblock{0.7965}{0.8324}{0.5902} & \metricblock{0.8029}{0.7719}{0.6009} & \metricblock{0.7911}{0.7915}{0.5756} & \metricblock{0.7620}{0.8124}{0.5192} \\
\addlinespace
RAUQ$_{\text{RO}}$ & \metricblock{0.8013}{0.7883}{0.5960} & \textbf{\metricblock{0.8037}{0.7483}{0.6134}} & \metricblock{0.7975}{0.8338}{0.5925} & \metricblock{0.8030}{0.7719}{0.6009} & \metricblock{0.7917}{0.7888}{0.5766} & \metricblock{0.7636}{0.8145}{0.5214} \\
\addlinespace
RAUQ$_{\text{RO}}^{\text{All}}$ & \textbf{\metricblock{0.8016}{0.7896}{0.5967}} & \metricblock{0.8036}{0.7495}{0.613} & \textbf{\metricblock{0.7991}{0.8352}{0.5959}} & \textbf{\metricblock{0.8047}{0.7737}{0.6042}} & \textbf{\metricblock{0.7953}{0.7943}{0.5842}} & \metricblock{0.7508}{0.8056}{0.4962} \\
\addlinespace
RAUQ$_{\text{M-H}}^{\text{All}}$ & \metricblock{0.8002}{0.788}{0.5967} & \metricblock{0.8030}{0.7475}{0.6128} & \metricblock{0.7968}{0.8331}{0.5910} & \metricblock{0.8030}{0.7718}{0.6013} & \metricblock{0.7918}{0.7920}{0.5772} & \metricblock{0.7661}{0.8152}{0.5264} \\
\bottomrule
\end{tabular}
\caption{Overall performance of all methods (rows) across all models (columns). Scores are stacked as AUROC / AURAC / PRR. The best AUROC per model is bold.
Model Abbreviations: LLaMA (LLaMA-3.1-8B-I), Mistral (Mistral-7B-I-v0.3), M-Nemo (Mistral-Nemo-I-2407), Falcon (Falcon-10B-I), Qwen (Qwen-2.5-7B-I), Gemma (Gemma-2-9B-It).
Method Abbreviations: ES (EigenScore), PPL (Perplexity), Sem./Norm./Pred. Ent. (Semantic/Normalized/Predictive Entropy), M-H (Mean Heads), RO (Rollout), In (Input), All (All Tokens).}
\label{tab:models}

\end{table}

\section{Conclusion}
This work tackled the challenge of hallucination detection in large language models by emphasizing the importance of distinguishing between intrinsic and extrinsic hallucinations. We used the RAUQ algorithm and introduced novel attention-based variants that are both more interpretable and more effective across various benchmarks. To rigorously evaluate these methods, we proposed a structured framework that separates hallucination types and aggregates scores over a concatenation of datasets, offering a more robust measure of performance.

Our findings reveal a clear trade-off among existing approaches. Despite the computational overhead, sampling-based methods, such as Semantic Entropy, are particularly effective for detecting extrinsic hallucinations, where the model lacks factual knowledge. In contrast, our proposed variants that aggregate attention over input tokens perform best on intrinsic hallucinations, which stem from contradictions or ambiguities in the input. These results underscore the need to align detection strategies with the nature of the underlying hallucination. Despite their simplicity, attention-based uncertainty methods offer an efficient and scalable alternative to sampling methods, while still achieving competitive or superior results, indicating a promising direction for hallucination detection in more sophisticated tasks.

\section*{Acknowledgments}
This publication was made possible by the use of the FactoryIA supercomputer, financially supported by the Ile-De-France Regional Council.

\small 
\bibliography{Main/LaTeX/bib}
\end{document}